\documentclass{article}
\usepackage{times}
\usepackage{algorithm,algorithmic,a4wide,amssymb}
\usepackage{graphicx}
\usepackage{stfloats}
\usepackage{fixltx2e}
\usepackage[caption=false,font=footnotesize]{subfig}
\graphicspath{{./figures/}}

\newcommand{\cf}{\emph{cf}.}

\begin{document}

\title{First Experiments with {\sc PowerPlay}}
\date{2012}
\author{Rupesh Kumar Srivastava, Bas R.~Steunebrink and J\"{u}rgen Schmidhuber \\
The Swiss AI Lab IDSIA, Galleria 2, 6928 Manno-Lugano \\
University of Lugano \& SUPSI, Switzerland}
\maketitle

\maketitle

\begin{abstract} Like a scientist or a playing child,  {\sc PowerPlay} \cite{Schmidhuber:11powerplay}
not only learns new skills to solve given problems, but also {\em invents} new interesting problems by itself.  
By design, it continually comes up
  with the fastest to find, initially novel, but eventually solvable tasks.
  It also continually simplifies or compresses or speeds up solutions to previous
  tasks.  Here we describe first experiments with {\sc PowerPlay}.  A
  self-delimiting recurrent neural network SLIM RNN \cite{Schmidhuber:12slimnn} is used as a general computational
problem solving  architecture.  Its connection  weights can encode arbitrary, self-delimiting, halting or
  non-halting programs affecting both environment (through effectors) and
  internal states encoding abstractions of event sequences.  Our {\sc PowerPlay}-driven SLIM RNN learns to become an  increasingly
  general solver of self-invented problems, continually adding new problem solving procedures to
  its growing skill repertoire. 
  Extending a recent conference paper \cite{rupesh2012icdl}, 
we identify interesting, emerging, developmental stages of our open-ended system.
We also show how it automatically self-modularizes, frequently re-using code for previously invented skills,
always trying to invent novel tasks that can be quickly validated because they do not require too 
many weight changes affecting too many previous tasks.
\end{abstract}

\section{Introduction} \label{intro}

 To automatically construct an increasingly general problem solver, the recent {\sc PowerPlay} framework
\cite{Schmidhuber:11powerplay} incrementally and efficiently searches the space of possible
pairs of (1) new task descriptions (from the set of all computable task descriptions), and (2) modifications of the current problem solver. 
The search continues until the first pair is discovered for which (i) the current solver cannot solve the new task, 
and (ii) the modified solver provably solves all previously learned tasks plus the new one. 
Here a new task may actually be to simplify, compress, or speed up previous solutions, which in turn may invoke or partially
re-use solutions to other tasks.
The above process of discovering and solving a novel task can be repeated forever in open-ended fashion.

As a concrete implementation of the solver, we use a special neural network (NN)  \cite{bishop:2006}
architecture  called the Self-Delimiting NN or SLIM NN \cite{Schmidhuber:12slimnn}.
Given a SLIM NN that can already solve a finite known set of previously learned tasks, 
an asymptotically 
optimal program search algorithm \cite{Levin:73,Schmidhuber:97bias,Schmidhuber:03nips,Schmidhuber:04oops} 
can be used to find a new pair that provably has properties (i) and (ii).
 Once such a pair is found, the cycle repeats itself. This results in a continually growing set of tasks solvable by an increasingly more powerful solver. The resulting repertoire of self-invented problem-solving procedures or skills can be exploited at any time to solve externally posed tasks.

The  SLIM NN has modifiable components, namely, its connection weights. By keeping
track of which tasks depend on each connection, {\sc PowerPlay} can reduce
the time required for testing previously solved tasks with certain newly modified connection weights, because only tasks
that depend on the changed connections need to be retested. If the solution of the most recently invented 
task does not require changes of many weights, and if the changed connections
do not affect many previous tasks, then validation may be very efficient. Since
{\sc PowerPlay}'s efficient search process has a built-in bias towards tasks whose validity check
requires little computational effort, there is an implicit incentive to generate
weight modifications that do not impact too many previous tasks. This leads to a natural
decomposition of the space of tasks and their solutions into more or less
independent regions. Thus, divide and conquer strategies are natural by-products
of {\sc PowerPlay}. 

Note that active learning methods \cite{Fedorov:72} such as AdaBoost \cite{Freund1997119} 
have a totally different set-up and purpose: there the user  
provides a set of samples to be learned, then each new classifier in a  
series of classifiers focuses on samples badly classified by previous  
classifiers. In open-ended {\sc PowerPlay}, however, all computational tasks  
(not necessarily classification tasks) can be self-invented; there is no  need for a
pre-defined global set of tasks that each new solver tries to solve  
better, instead the task set continually grows based on which task is  
easy to invent and validate, given what is already known.

Unlike our first implementations of curious~/ creative~/ playful agents from the
1990s \cite{Schmidhuber:91singaporecur,Storck:95,Schmidhuber:99cec} (\cf~\cite{Barto:12intrinsic,Dayan:12intrinsic,Oudeyer:12intrinsic,Nemzhov:12intrinsic}),
{\sc PowerPlay} provably (by design) does not have any problems with online
learning---it cannot forget previously learned skills, automatically segmenting
its life into a sequence of clearly identified tasks with explicitly recorded
solutions.  Unlike the task search of {\em theoretically optimal} creative agents
\cite{Schmidhuber:06cs,Schmidhuber:10ieeetamd}, {\sc PowerPlay}'s task search is
greedy, yet practically feasible. Here we present first experiments,
extending recent work \cite{rupesh2012icdl}.

\section{Notation \& Algorithmic Framework for {\sc PowerPlay} (Variant II)} \label{framework}

We use the notation of the original paper \cite{Schmidhuber:11powerplay},
and briefly review the basics relevant here.
$B^*$ denotes the set of finite bit strings over the binary alphabet
$B=\{0,1\}$,
$\mathbb{N}$ the natural numbers, $\mathbb{R}$ the real numbers.
The computational architecture of {\sc PowerPlay}'s problem solver may be a deterministic
universal computer, or a more limited device such as 
a feedforward NN.  All problem solvers
can be uniquely encoded \cite{Goedel:31} or implemented on universal computers
such as universal Turing Machines (TM) \cite{Turing:36}. Therefore, without loss
of generality, we can assume a fixed universal reference
computer whose inputs and outputs are elements of $B^*$.  User-defined
subsets ${\cal S}, {\cal T} \subset B^*$ define the sets of possible problem solvers
and task descriptions. For example,
$\cal T$ may be the infinite set of all computable tasks,
or a small subset thereof.
 ${\cal P} \subset B^*$ defines a set of possible programs
which may be used to generate or modify members of $\cal S$ or $\cal T$.
If our solver is a feedforward NN, then $\cal S$
could be a highly restricted subset of programs encoding the NN's possible
topologies and weights,
$\cal T$ could be encodings of input-output pairs for a supervised learning 
task, and $\cal P$ could be an algorithm that modifies the weights of the 
network.

The problem solver's initial program is called $s_0$.  
A particular sequence of
unique task descriptions $T_1, T_2, \ldots$ (where each $T_i \in {\cal T}$)
is chosen or ``invented'' by a search method (see below) such
that the solutions of  $T_1, \ldots, T_i$ can be computed by $s_i$, the
$i$-th instance of the program, but $T_i$ cannot be solved by $s_{i-1}$.  Each
$T_i$ consists of a unique problem identifier that can be read by  $s_i$ through
some built-in mechanism (e.g.,  input neurons of an NN as in
Sec.~\ref{experiment1} and \ref{experiment2}), and a unique description of a
deterministic procedure for deciding whether the problem has been solved.
For example, a simple task may require the solver to answer a particular input
pattern with a particular output pattern.  Or it may require the solver to steer
a robot towards a goal through a sequence of actions.  Denote $T_{\leq i}=
\{T_1, \ldots, T_i\}$; $T_{< i} = \{T_1, \ldots, T_{i-1}\}$. A valid task 
$T_i$ ($i>1$) may require solving at least one previously solved task
$T_k$ ($k<i$) more efficiently, by using less resources such as storage space,
computation time, energy, etc.\ quantified by the function $Cost(s, T)$. The algorithmic
framework (Alg.~\ref{cost}) incrementally trains the problem solver by finding
$p \in {\cal P}$ that increase the set of solvable tasks. For more details,
the reader is encouraged to refer to the  original report \cite{Schmidhuber:11powerplay}.

\pagebreak

\begin{algorithm}
\small
  \caption{{\sc PowerPlay} Framework (Variant II)}
  \label{cost}
  \begin{algorithmic}
    \STATE Initialize $s_0$ in some way 
    \FOR {$i := 1, 2, \ldots$} 
      \STATE Declare new global variables $T_i \in {\cal T}$,  $s_i \in {\cal S}$,  
      $p_i  \in {\cal P}$, $c_i, c^*_i \in \mathbb{R}$ (all unassigned)
      \REPEAT 
      \STATE Let a search algorithm (e.g.,
      Section \ref{experiment1}) set $p_i$, a new candidate program. Give $p_i$
      limited time to do:
      \STATE  {\bf *} {\sc Task Invention}: Unless the user specifies  $T_i$, let $p_i$ set  $T_i$. 
      \STATE  {\bf *}  {\sc Solver Modification}: Let $p_i$ set $s_i$  by computing a
      modification of $s_{i-1}$. 
      \STATE   {\bf *}  {\sc Correctness Demonstration}: Let $p_i$  compute $c_i := Cost(s_i, T_{\leq i})$ and $c^*_i
      := Cost(s_{i-1}, T_{\leq i})$ \UNTIL {$c^*_i - c_i > \epsilon$ (minimal
      savings of costs such as time/space/etc on all tasks so far)}
      \STATE  Freeze/store forever $p_i, T_i, s_i, c_i, c^*_i$
    \ENDFOR
  \end{algorithmic}
\end{algorithm}

\section{Experiment 1: Self-Invented Pattern Recognition Tasks}
\label{experiment1}
We start with 
pattern classification tasks. In this setup, $s$ encodes an arbitrary set of
weights for a fixed-topology multi-layer perceptron (MLP).
The MLP
maps two-dimensional, real-valued input vectors from the unit square to binary
labels; i.e., $s$: $[0,1)\times[0,1) \rightarrow {0,1}$. The output label is 0 or 1 depending
on whether or not the real-valued activation of the MLP's single output
neuron exceeds 0.5. Binary programs $p \in {\cal P}$ 
of length $\mathit{length}(p)$
compute tasks and modify
$s$ as follows. If $p^1$ (the first bit of $p$) is 0, this will specify that the
current task is to simplify $s$ by weight decay, under the assumption that
smaller weights are simpler. Such programs implement compression tasks. But if $p^1$ is 1, then the target label of the
current task candidate $T$ will be given by the next bit $p^2$, and $T$'s
two-dimensional input vector will be uniquely encoded by the remainder of $p$'s bit string, $p^3p^4 \ldots p^n$, as follows. The string $p^3p^4 \ldots p^n$ is taken as the binary representation of an integer $N$. Then a 2D Gaussian pseudo-random number generator is used to generate numbers $(x_1, y_1), (x_2, y_2), \ldots$, where $x$ and $y$ are used as 2D coordinates in the unit square. Now the task is to label the coordinates $(x_N, y_N)$ as $p_2$.

The random number generator is re-seeded by the same seed
every time a new task search 
begins, thus ensuring a deterministic search order.
Since we only have two labels in this experiment, we do not need $p^2$ as we 
can choose the target label to be different from the label currently 
assigned by the MLP to the encoded input.
To run $p$
for $t$ steps (on a training set of $i$ patterns so far) means to execute
$\lfloor t/2i \rfloor$  epochs of gradient descent on the
training set and check whether the patterns are correctly classified. Here one step always 
refers to the processing of a single pattern (either a forward or 
backward pass), regardless of the task.

Assume now that {\sc PowerPlay} has already learned a version of $s$ called $s_{i-1}$
able to classify $i-1$ previously invented training patterns ($i > 1$). Then the next
task is defined by a simple enumerative search in the style of universal
search \cite{Levin:74,Schmidhuber:97bias,Schmidhuber:04oops}, 
which combines task simplification and systematic run-time growth (see
Alg.~\ref{backprop}).

\begin{algorithm} 
\small
  \caption{{\sc PowerPlay} implementation for experiment 1}
 \label{backprop}
 \begin{algorithmic}
   \STATE {Initialize $s_0$ in some way}
 \FOR {$i := 1, 2, \ldots$}
 \FOR {$m := 1, 2, \ldots$}
 \FORALL {candidate programs $p$ s.t. $\mathit{length}(p) \leq m$}
 	\STATE Run $p$ for at most $2^{m-\mathit{length}(p)}$ steps
	\IF {$p$ creates $s_{i}$ from $s_{i-1}$ correctly classifying all $i$ training 
	patterns so far \AND ($s_{i}$ is substantially simpler than 
	$s_{i-1}$
        \OR $s_{i}$ can classify
	a newly found pattern misclassified by $s_{i-1}$)}
        \STATE {Set $p_i$ := $p$ (store the candidate)}
        \STATE {\bf exit $m$ loop};
	\ENDIF
      \ENDFOR
\ENDFOR
\ENDFOR
  \end{algorithmic}
\end{algorithm} 

\begin{figure*}[!t]
  \centering
  \subfloat[After 1 task]{\includegraphics[width=0.33\linewidth]{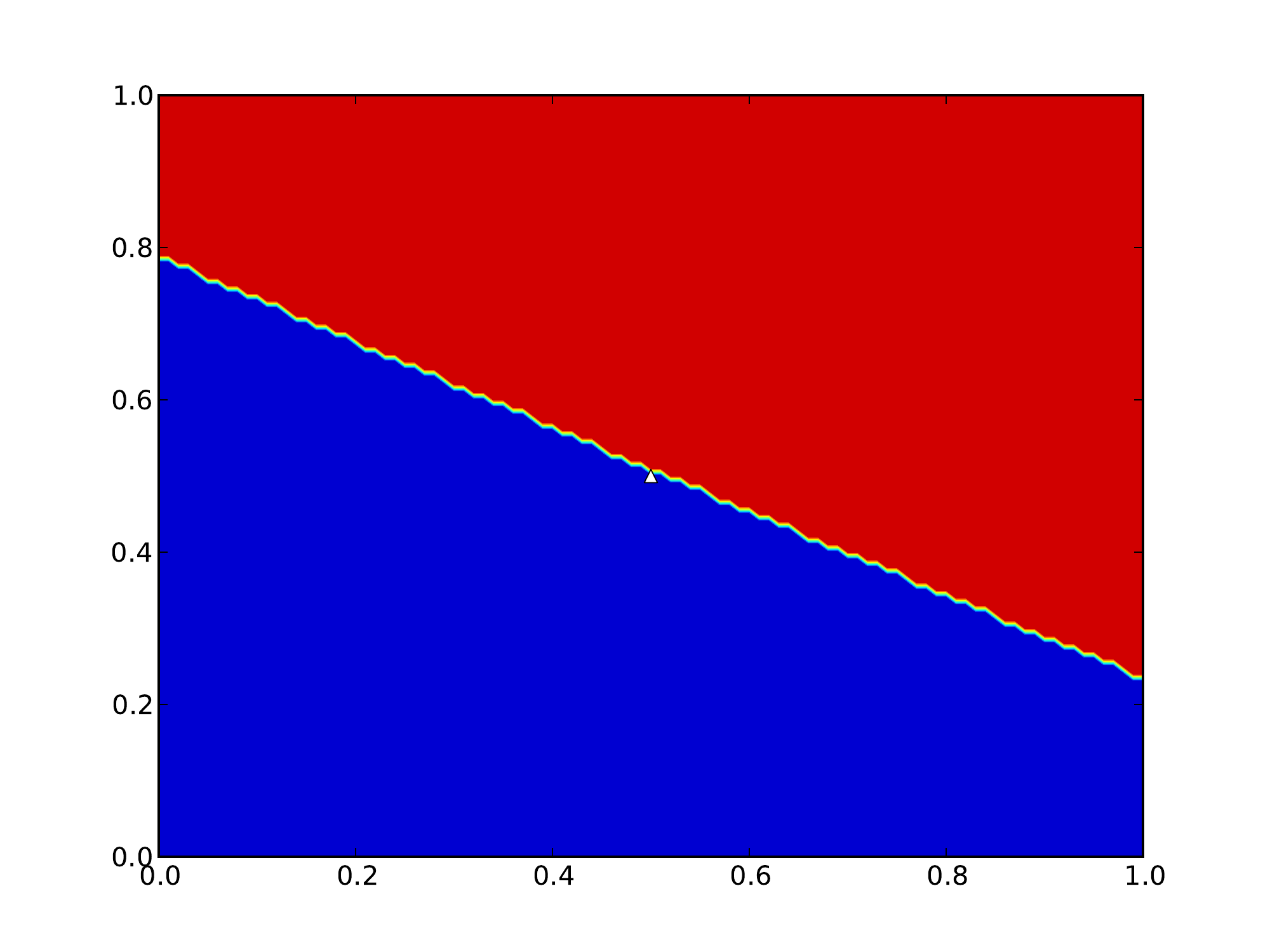}}
  \subfloat[After 16 tasks]{\includegraphics[width=0.33\linewidth]{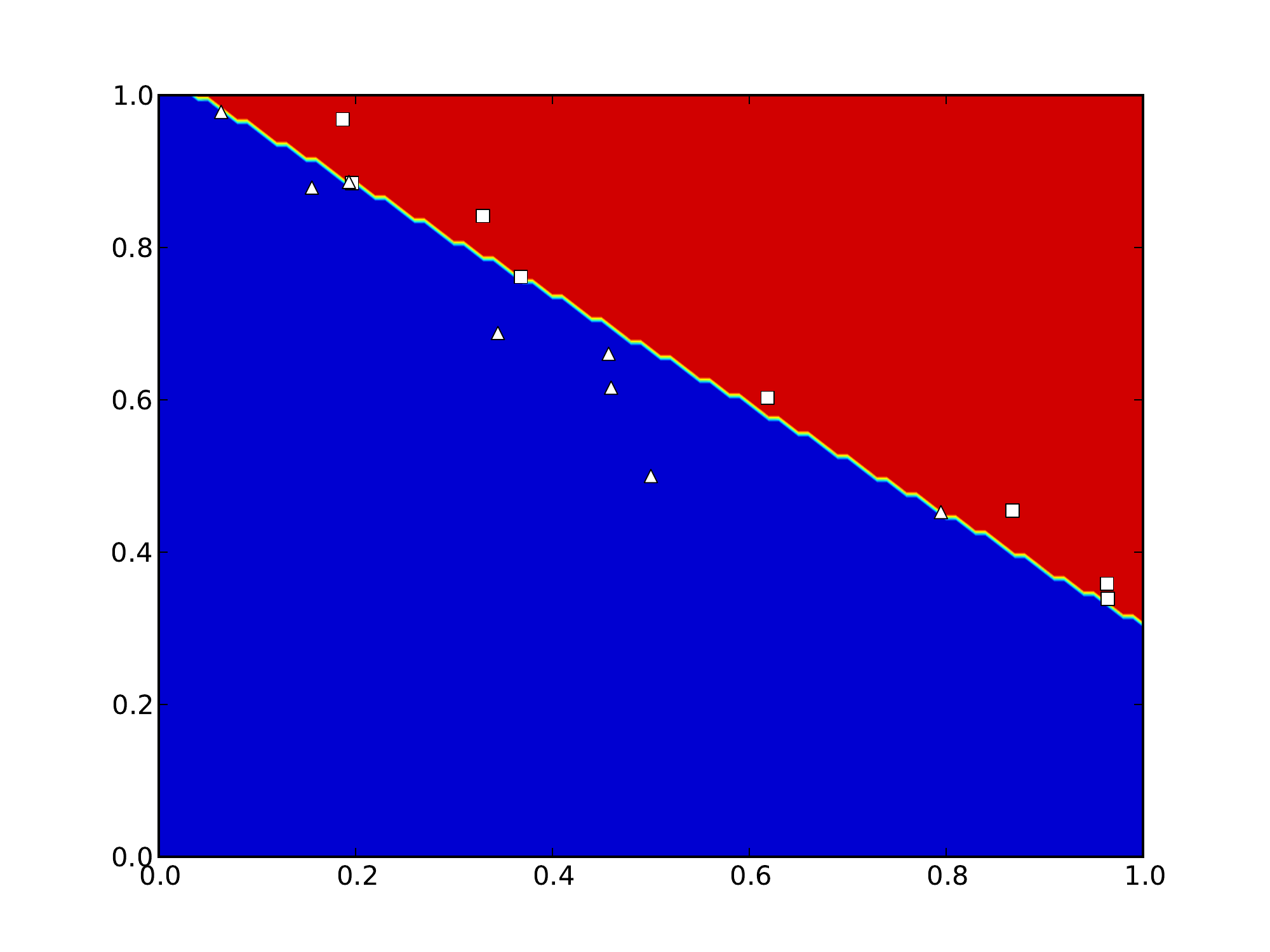}}
  \subfloat[After 17 tasks]{\includegraphics[width=0.33\linewidth]{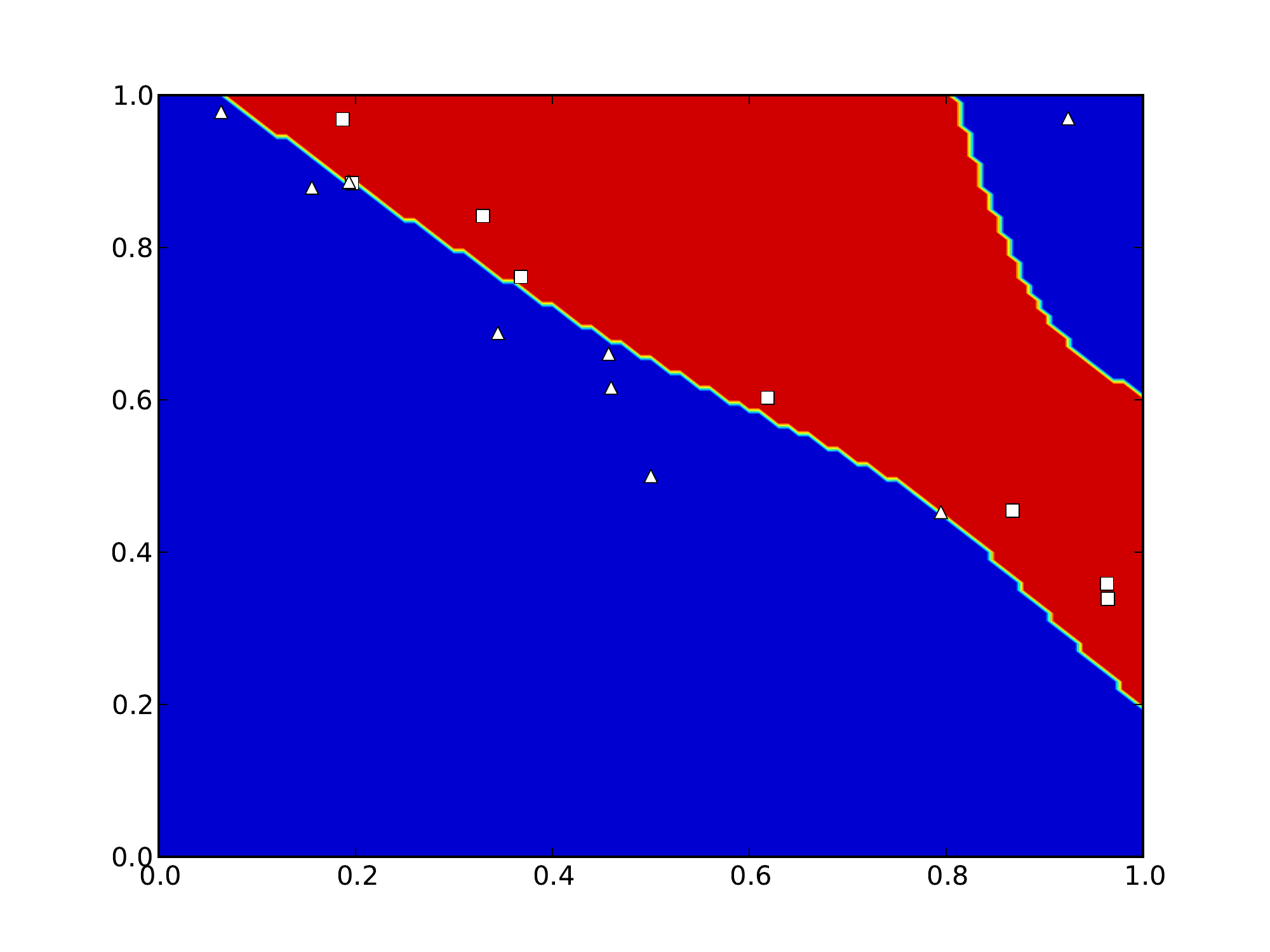}}
\\
  \subfloat[After 25 tasks]{\includegraphics[width=0.33\linewidth]{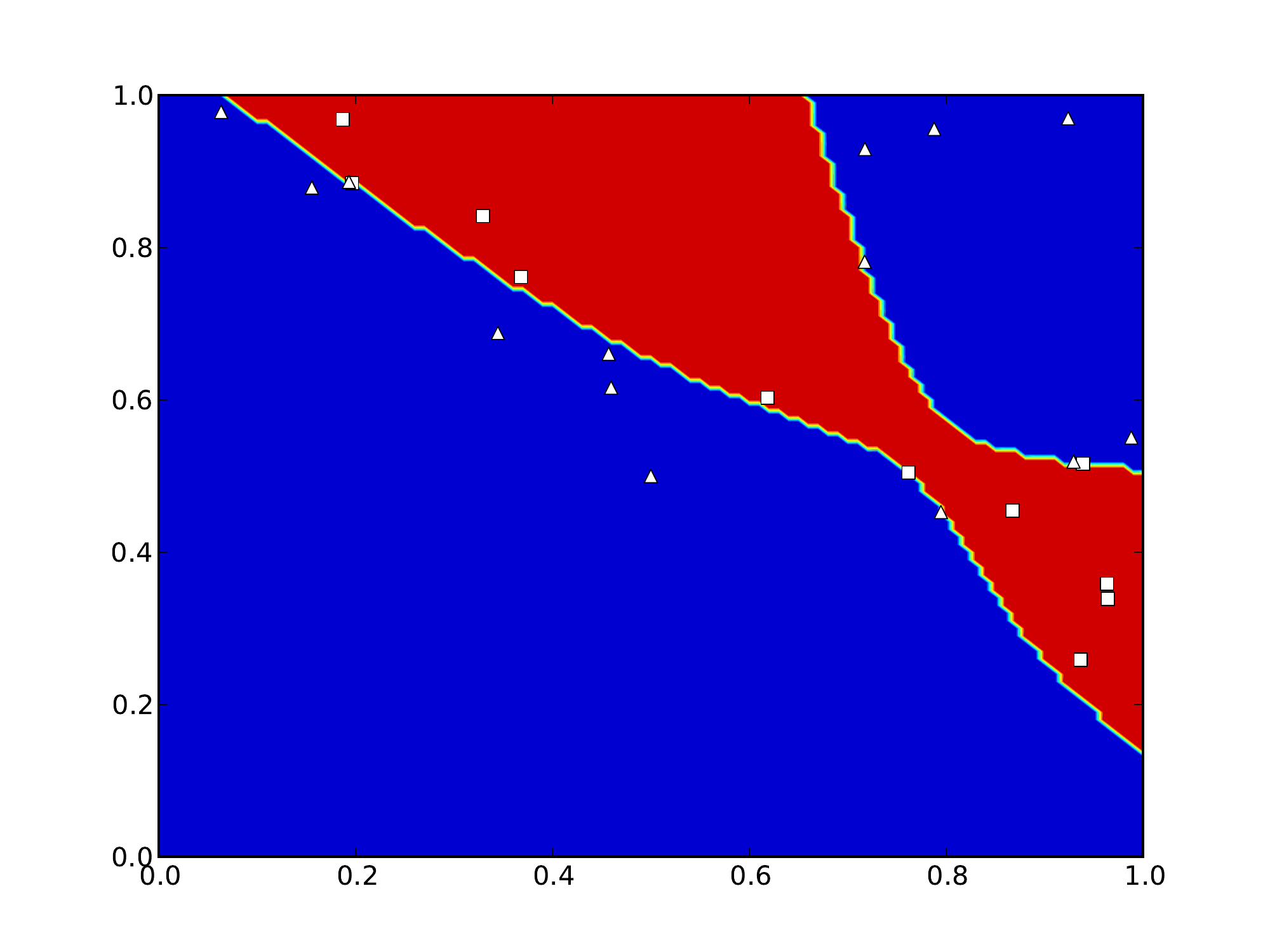}}
  \subfloat[After 33 tasks]{\includegraphics[width=0.33\linewidth]{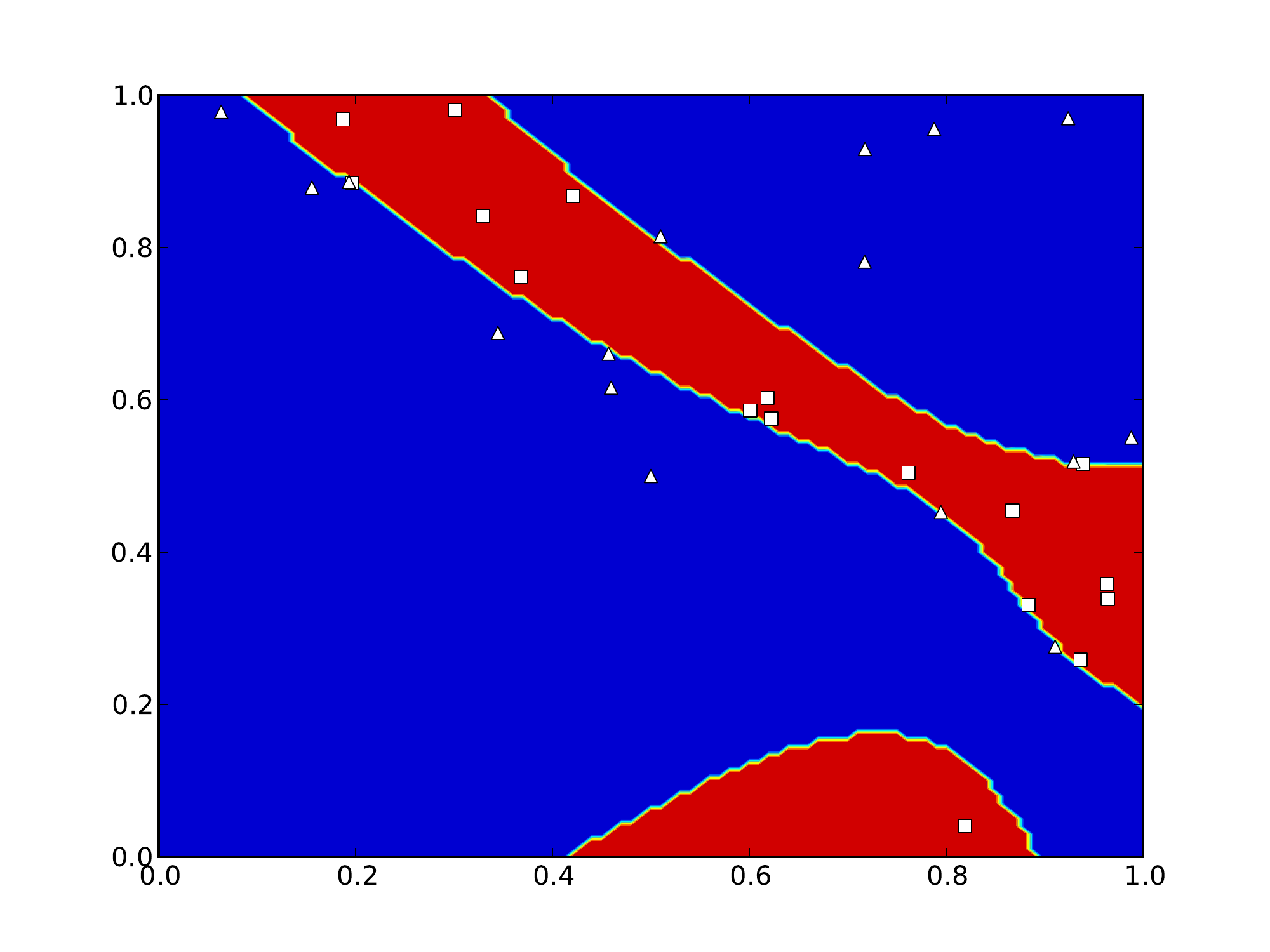}}
  \subfloat[After 43 tasks]{\includegraphics[width=0.33\linewidth]{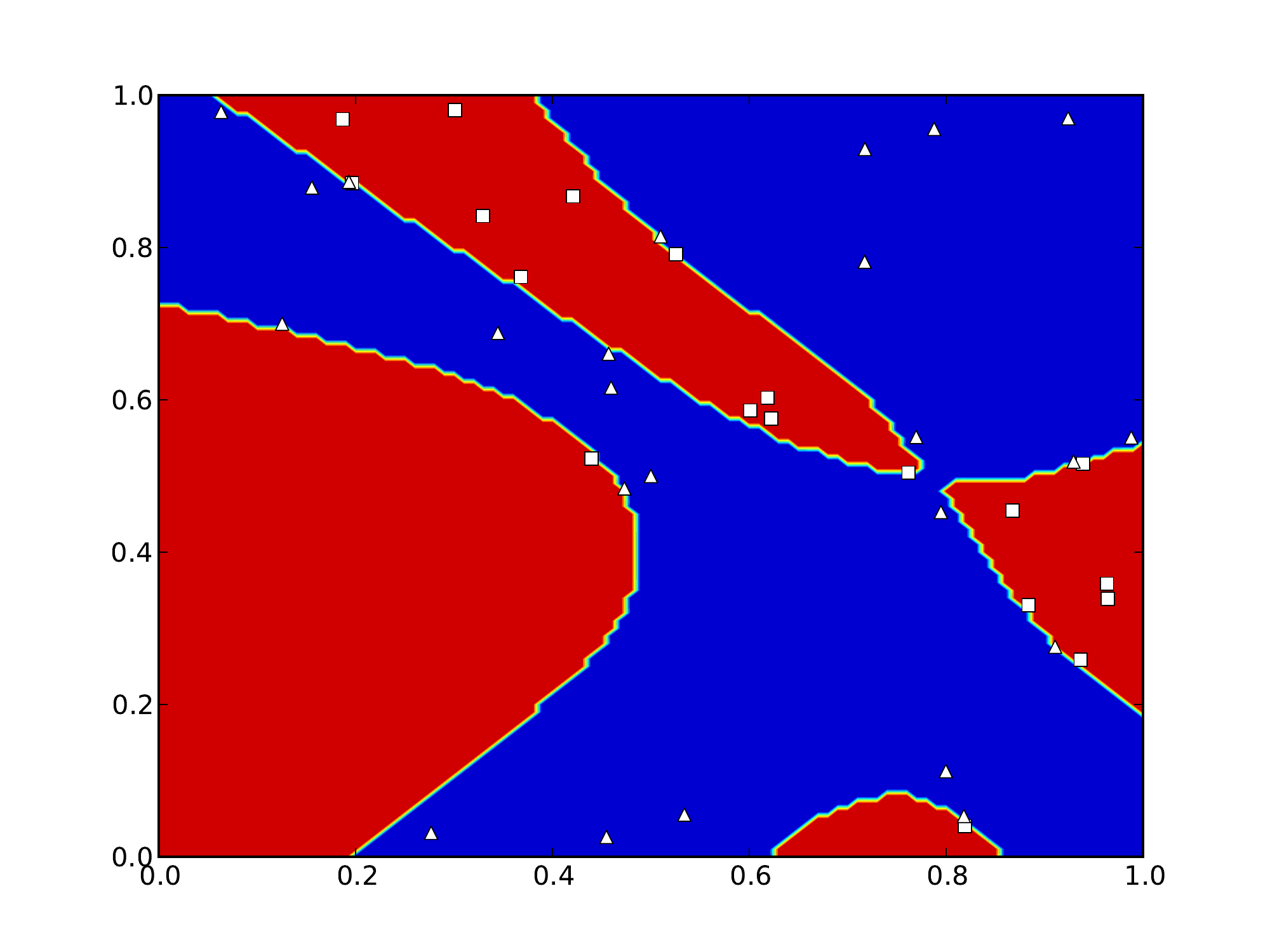}}
  \caption{ 
Experiment 1. Right after initialization, before the first compressions, 
the decision boundary may be arbitrary and possibly non-linear.
The drive to compress and simplify, however, first encourages linear separability (top row).  
As more associations are invented, it becomes
harder and harder to learn new ones that break the previous solver's generalization ability,
while maintaining a linear boundary.
Eventually this causes the decision boundary to become non-linear (bottom row). 
The decision boundary becomes increasingly non-linear, as more and more associations are invented and learned.}
  \label{fig:exp1:generalization}
\end{figure*}

Since the compression task code is the single bit `0',
roughly half of the total search time is spent on simplification,
the rest is spent on the invention of new training patterns that break the
MLP's current generalization ability.

To monitor the evolution of the solver's
{\em generalization map},
after each successful search for a new task, the labels of grid points are
plotted in a rather dense grid on the unit square (Fig.~\ref{fig:exp1:generalization}), to see how
the MLP maps $[0,1)\times[0,1)$ to ${0,1}$.
As expected, the experiments show that in the beginning
{\sc PowerPlay} prefers to invent and learn simple linear functions.
However, there is a phase transition to more
complex non-linear functions after a few tasks, 
indicating a new developmental stage
\cite{Piaget:55,Schmidhuber:02predictable,Hung2011}.
This is a natural by-product of the search for simple tasks---they are easier to invent and verify than more complex non-linear tasks.
As learning proceeds, we observe 
that the decision boundary becomes increasingly non-linear, because the system has to come up with tasks which the solver cannot solve yet, but the solver becomes increasingly more powerful, so the system has to invent increasingly harder tasks. 
On the other hand, the search time for solutions to
harder and harder tasks need not grow over time, since new solutions do not have to be learnt from scratch,
but may re-use previous solutions encoded as parts of the previous solver.

\section{Experiment 2: Self-Invented Tasks Involving Motor Control and Internal
Abstractions} \label{experiment2}

\subsection{Self-Delimiting (SLIM) Programs Run on A Recurrent Neural Network (RNN)}
\label{rnn}
Here we describe experiments with a {\sc PowerPlay}-based RNN that continually invents novel 
sequences of actions affecting an external environment, over time becoming a more and more general solver 
of self-invented problems. 

RNNs are general computers that allow for both sequential and
parallel computations.  Given enough neurons and an appropriate weight matrix,
an RNN can compute any function computable by a standard PC
\cite{Schmidhuber:90thesis}.
We use a particular RNN named SLIM RNN \cite{Schmidhuber:12slimnn} to define 
$\cal S$ for our experiment. 
Here we briefly review its basics.

The $k$-th computational unit or {\em neuron} of our SLIM RNN is denoted $u^k$
($0<k \leq n(u) \in \mathbb{N}$).
 $w^{lk}$ is the real-valued {\em weight} on the directed connection  $c^{lk}$
from $u^l$ to $u^k$. 
 At discrete time step $t=1,2,\ldots,t_{end}$ of a
finite interaction sequence with the environment, $u^k(t)$ denotes the
real-valued {\em activation} of $u^k$.  
There are designated neurons serving as \emph{online inputs}, which read 
real-valued observations from the environment, and \emph{outputs} whose 
activations encode actions in the environment, e.g.,
the movement commands for a robot. We initialize all $u^k(1)$ = 0 and
compute $u^k(t+1)=f^k(\sum_l w^{lk}u^l(t))$ where $f$ may be of the form
$f^k(x)=1/(1+e^{-x})$, or $f^k(x)=x$, or $f^k(x)=1$ if $x \geq 0$ and 0
otherwise. To program the SLIM RNN means to set the weight matrix $\langle
w^{lk} \rangle$.

A special feature of the SLIM RNN is that it has a single \emph{halt} 
neuron with a fixed \emph{halt-threshold}. If at 
any time $t$ its activation exceeds the \emph{halt-threshold},
the network's computation stops. Thus, any network topology in 
which there exists a path from the online or task inputs to the halt neuron
can run self-delimiting programs \cite{Levin:74,Chaitin:75,Schmidhuber:97bias,Schmidhuber:04oops}
studied in the theory of Kolmogorov complexity and algorithmic probability
\cite{Solomonoff:64,Kolmogorov:65}.
Inspired by a previous architecture \cite{Schmidhuber:89cs}, neurons other than the inputs and 
outputs in our RNN are arranged in winner-take-all subsets
(WITAS) of $n_{\mathit{witas}}$ neurons each ($n_{\mathit{witas}} = 4$ was used for this experiment).
At each time step $t$, $u^k(t)$ is set to 1 if $u^k$ is a  winning neuron in some WITAS (the one 
with the highest activation), and  to 0 otherwise.
This feature gives the SLIM RNN the potential to modularize itself, since 
neurons can act as \emph{gates} to various self-determined regions of the network. By regulating 
the information flow, the network may use only a fraction 
of the weights $\langle w^{lk} \rangle$ for a given task.

\begin{figure*}[t]
  \centering
  \subfloat[$t$ = 1]{\includegraphics[width=0.45\linewidth]{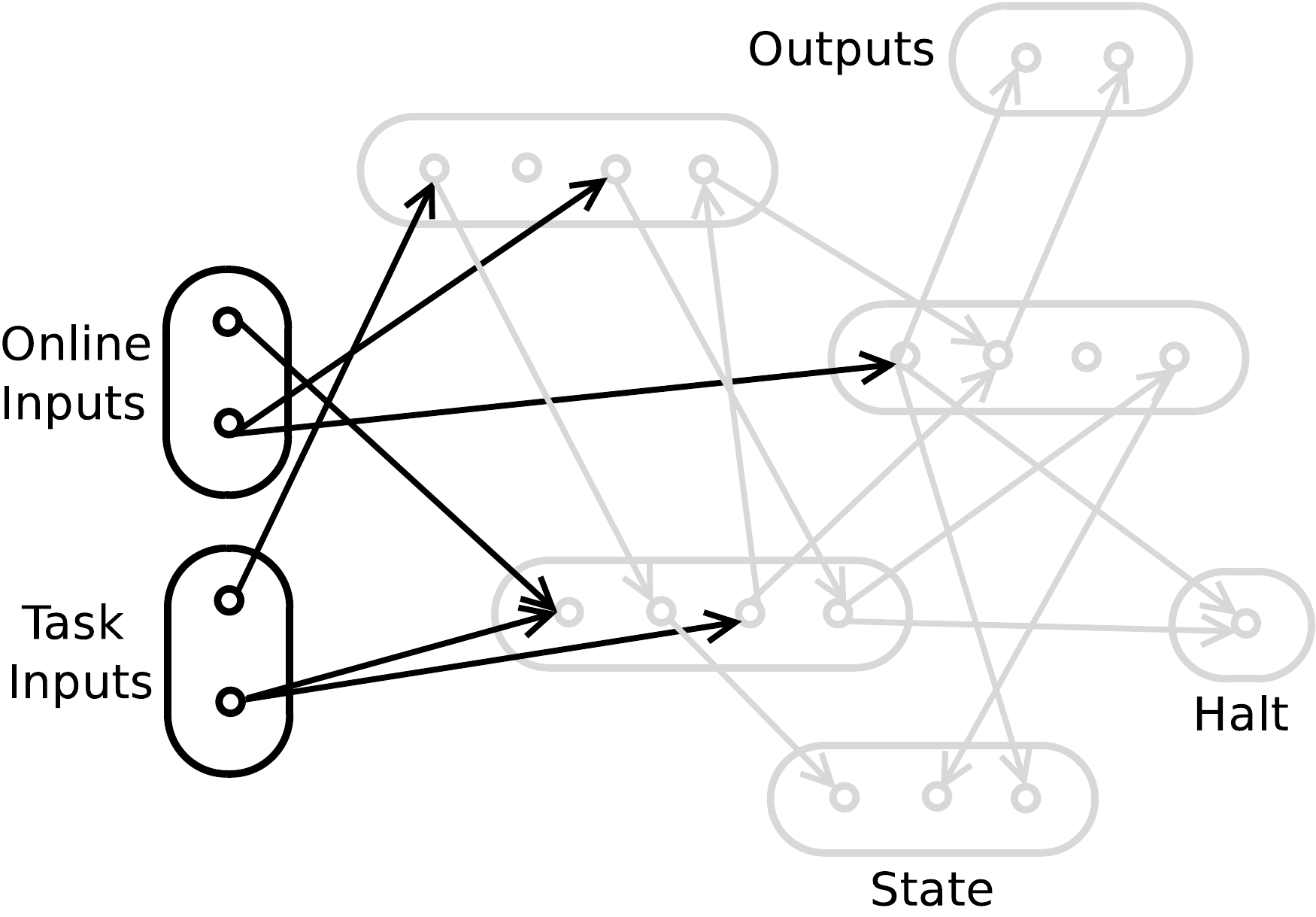}}
  \hspace{.04\linewidth}
  \subfloat[$t$ = 2]{\includegraphics[width=0.45\linewidth]{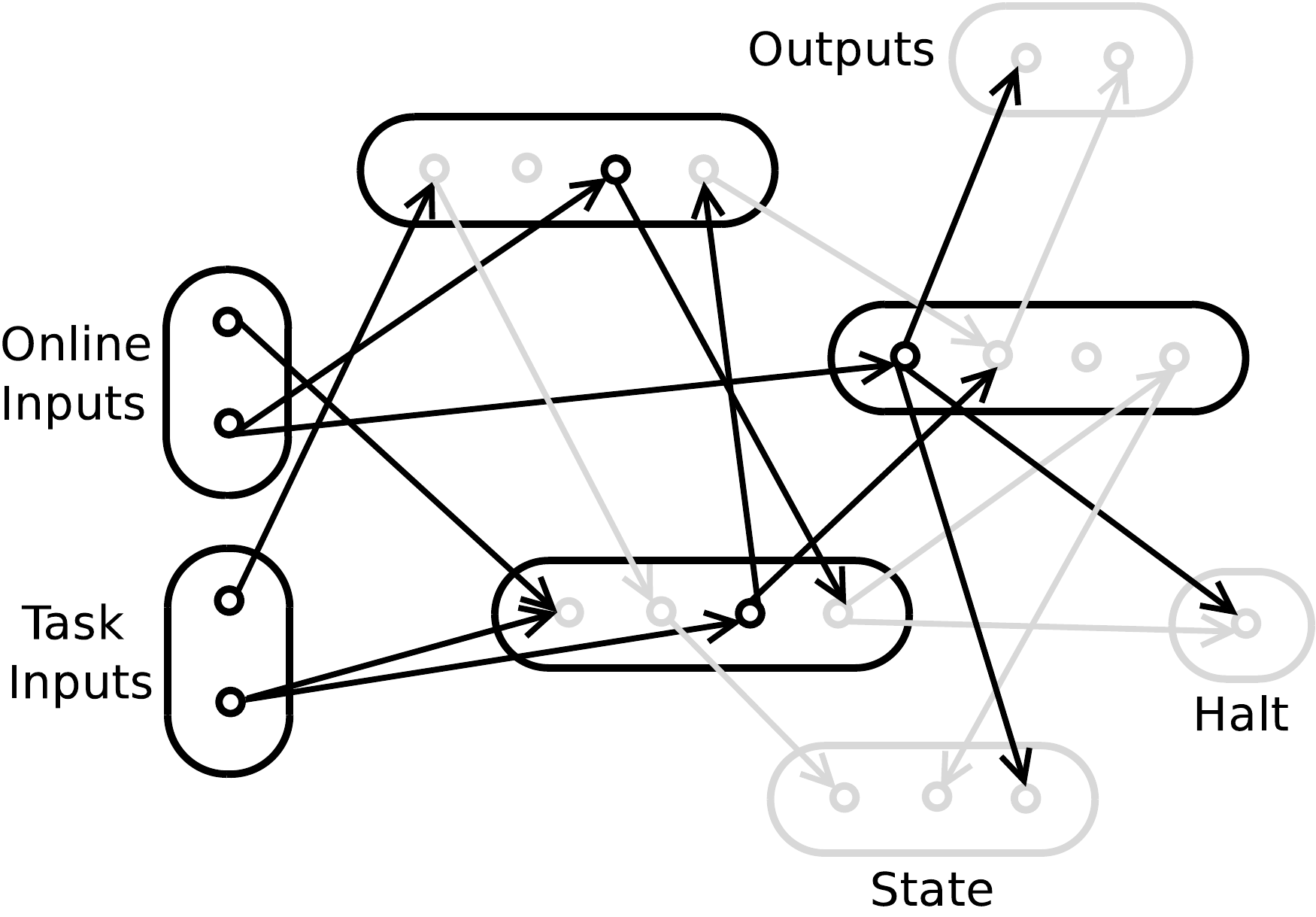}}
  \\
  \subfloat[$t$ = 3]{\includegraphics[width=0.45\linewidth]{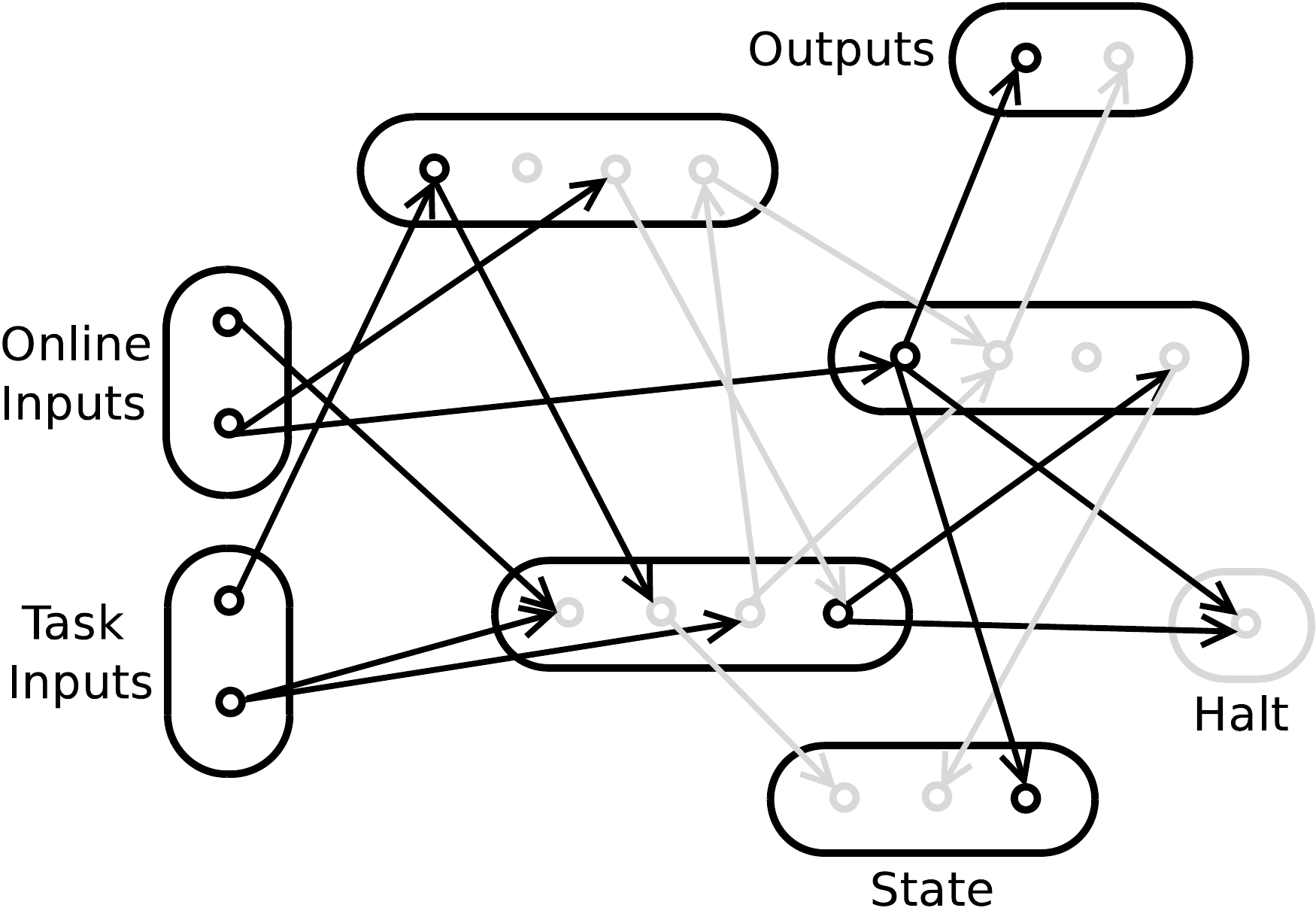}}
  \hspace{.04\linewidth}
  \subfloat[$t$ = 4]{\includegraphics[width=0.45\linewidth]{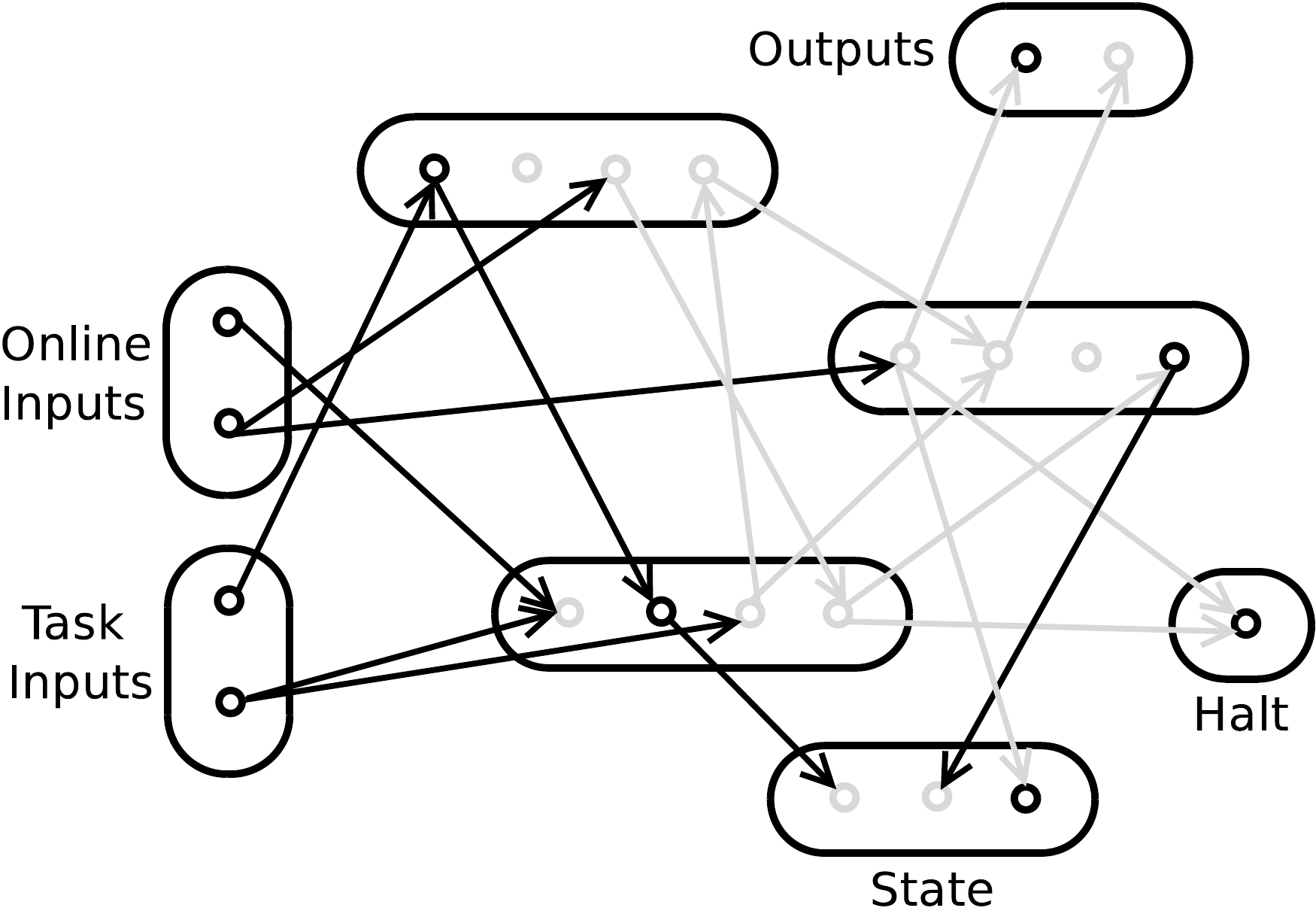}}
  \caption{SLIM RNN activation scheme. At various time steps, active/winning neurons  and their outgoing connections are highlighted. At each step, 
at most one neuron per WITAS can become active and  propagate activations through its outgoing connections.}
  \label{fig:rnnschematic}
\end{figure*}

Apart from the online input, output and halt neurons, 
a fixed number $n_{ti}$ of neurons are set to be \emph{task inputs}. These
inputs remain constant for $1 \leq t < t_{end}$ and serve as self-generated task specifications. 
Finally, there is a subset of $n_s$ 
 \emph{internal state} neurons whose activations are considered as the final outcome when the program halts.
Thus a \emph{non-compression} task is:
Given a particular task input, interact with the environment (read online inputs, 
produce outputs) until the network halts and produces a particular internal state---the abstract {\em goal}---which is read from the internal state neurons.
Since the SLIM RNN is a general computer, it can represent essentially
arbitrary computable tasks in this way.
Fig.~\ref{fig:rnnschematic} illustrates the network's activation 
spreading for a particular task.
A more detailed discussion of SLIM RNNs and their efficient implementation 
can be found in the original report  \cite{Schmidhuber:12slimnn}.

The SLIM RNN is trained on the fovea environment described in Sec.~\ref{fovea} 
using the {\sc PowerPlay} framework according to Algorithm~\ref{fovea-task} below. The difference to Algorithm~\ref{backprop} lies in task set-specific details such as the encoding of task 
inputs and the definition of `inventing and learning' a task. The bit string $p$ now 
encodes a set of $n_{ti}$ real numbers between 0 and 1 which denote the constant
task inputs for this program. 
Given a new set of task inputs, the new task is considered learned if the network halts and reaches a particular internal state (potentially after interacting with 
the environment), and remains able to properly 
reproduce the saved internal states for all previously
learned tasks. This is implemented by first checking if the network can halt and produce an internal state on the newly generated task inputs. Only if the network cannot halt within a chosen fraction of the time budget
dictated by $\mathit{length}(p)$, the length of program $p$,
the remaining budget is used for trying to learn the task using a simple mutation rule, by modifying a few weights of the network.
When $p$ is the single bit `0', the task is interpreted as a \emph{compression} task. Here compression either means a reduction of the sum of squared weights without increasing the total number of connection usages by all previously learned tasks, or a reduction of the total number of connection usages on all previously learned tasks without increasing the sum of squared weights.

\begin{algorithm} 
\small
  \caption{{\sc PowerPlay} implementation for experiment 2}
 \label{fovea-task}
 \begin{algorithmic}
   \STATE {Initialize $s_0$ in some way}
 \FOR {$i := 1, 2, \ldots$}
 \FOR {$m := 1, 2, \ldots$}
  \FORALL {candidate programs $p$ s.t. $\mathit{length}(p) \leq m$}
    \STATE Set $\mathit{time\_budget} := 2^{m-\mathit{length}(p)}$ 
  	\IF {$p$ encodes a compression task}
		\STATE Set $s_\mathit{temp} := s_{i-1}$
  		\WHILE {$\mathit{time\_budget} > 0$}
    		\STATE Create $s_i$ from $s_\mathit{temp}$ through random perturbation of a few connection weights
    		\IF {compression is successful \AND $\mathit{time\_budget} \geq 0$}
    			\STATE Set $s_\mathit{temp} := s_i$
    		\ENDIF
    	\ENDWHILE
    \ELSE		
  		\WHILE {$\mathit{time\_budget} > 0$}
     		\STATE Create $s_{i}$ from $s_{i-1}$ through random perturbation of a few connection weights
     	\STATE From $p$ generate task $k$
	 	\IF {$s_{i-1}$ does not solve $k$ \AND $s_{i}$ solves $k$ \AND $s_{i}$ solves all previous tasks in the repertoire \AND $\mathit{time\_budget} \geq 0$}
	 		\STATE Add the pair ($k$, internal state) to the repertoire
	 		\STATE {\bf exit $m$ loop}
		\ENDIF
   		\ENDWHILE
   	\ENDIF
  \ENDFOR
\ENDFOR
\ENDFOR
  \end{algorithmic}
\end{algorithm}

Since our {\sc Powerplay} variant methodically increases search time, half of which 
is used for compression, it automatically 
encourages the network to invent novel tasks 
that do not require many changes of weights used by many previous tasks.

Our SLIM RNN implementation efficiently resets activations computed by the numerous unsuccessful tested candidate
programs. We keep track of used connections and active (winner) neurons 
at each time step, to reset activations such that tracking/undoing effects 
of programs essentially does not cost more than their execution.

\subsection{RNN-Controlled Fovea Environment} \label{fovea}

\begin{figure}[t]
  \centering
  \vspace{-1ex}
  \subfloat[]{\includegraphics[width=0.3\linewidth]{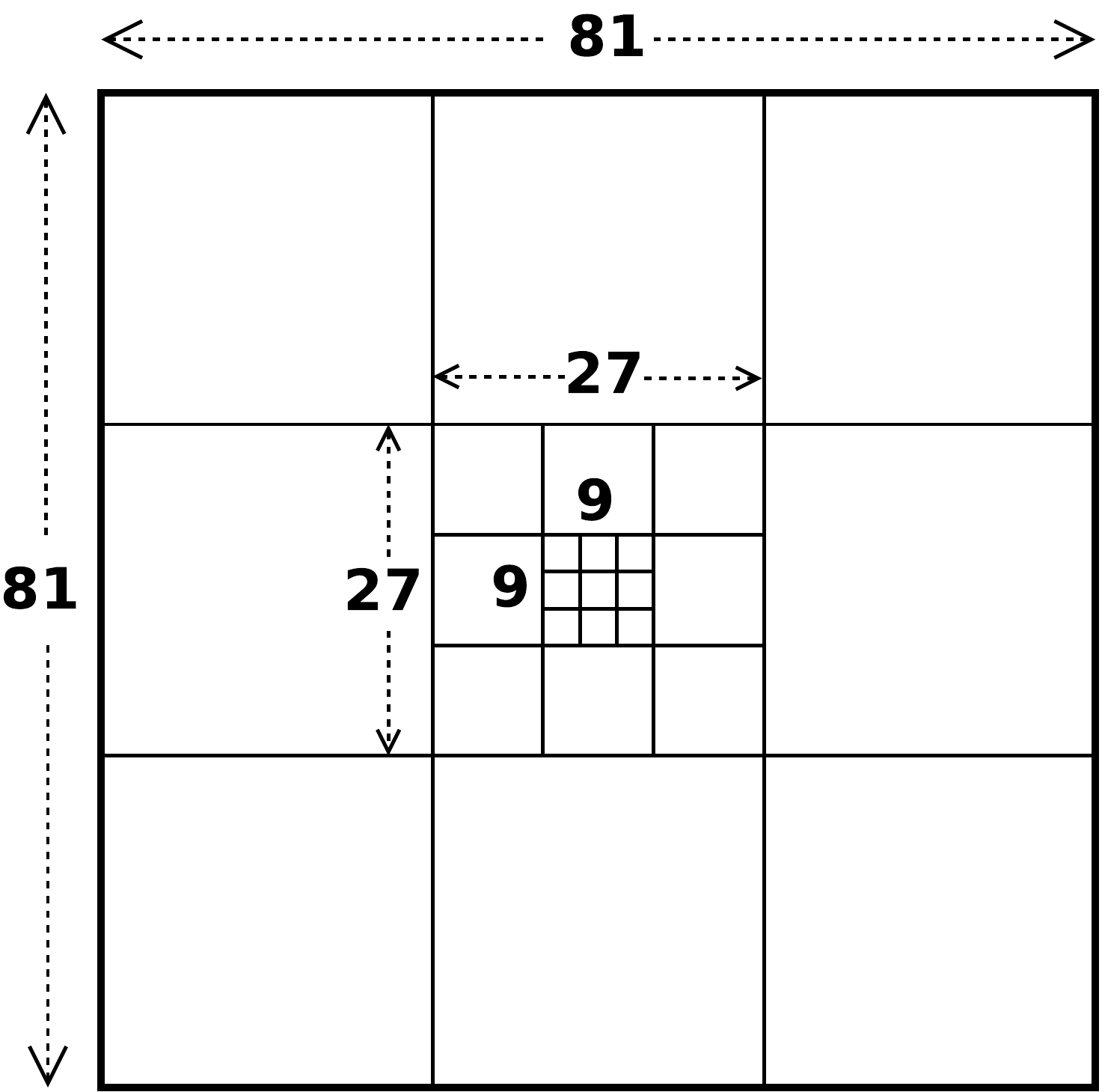}}
  \hspace{0.05\linewidth}
  \subfloat[]{\includegraphics[width=0.6\linewidth]{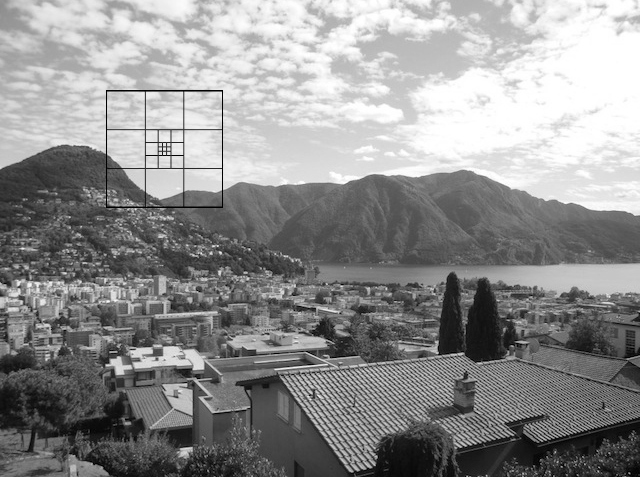}}
  \caption{(a) Fovea design. Pixel intensities over each square are averaged to produce a real valued input. The
    smallest squares in the center are of size $3 \times 3$.
    (b) The RNN controls the fovea movement over a static image, in our experiments this photo of the city of Lugano.}
  \label{figfoveaschematic}
\end{figure}

The environment for this experiment consists of a static image which is observed sequentially by the 
RNN through a fovea, whose movement it can control at each time step. The size of the fovea is 
$81 \times 81$ pixels; it produces 25 real valued online inputs (normalized to $[0, 1]$) by averaging the 
pixel intensities over regions of varying sizes such that it has higher resolution at the center 
and lower resolution in the periphery (Fig. \ref{figfoveaschematic}). The fovea is controlled 
using 8 real-valued outputs of the network, and a parameter \emph{win-threshold}. Out of the first four 
outputs, the one with the highest value greater than  \emph{win-threshold} is interpreted as 
a movement command: up, down, left, or right. If none of the first four outputs exceeds the 
threshold, the fovea does not move. Similarily, the next four outputs are interpreted as the 
fovea step size on the image (3, 9, 27 or 81 pixels in case of exceeding 
the threshold, 1 pixel otherwise).

\subsection{Results} \label{exp2:results}

\begin{figure*}[t]
  \centering
  \includegraphics[width=0.8\linewidth]{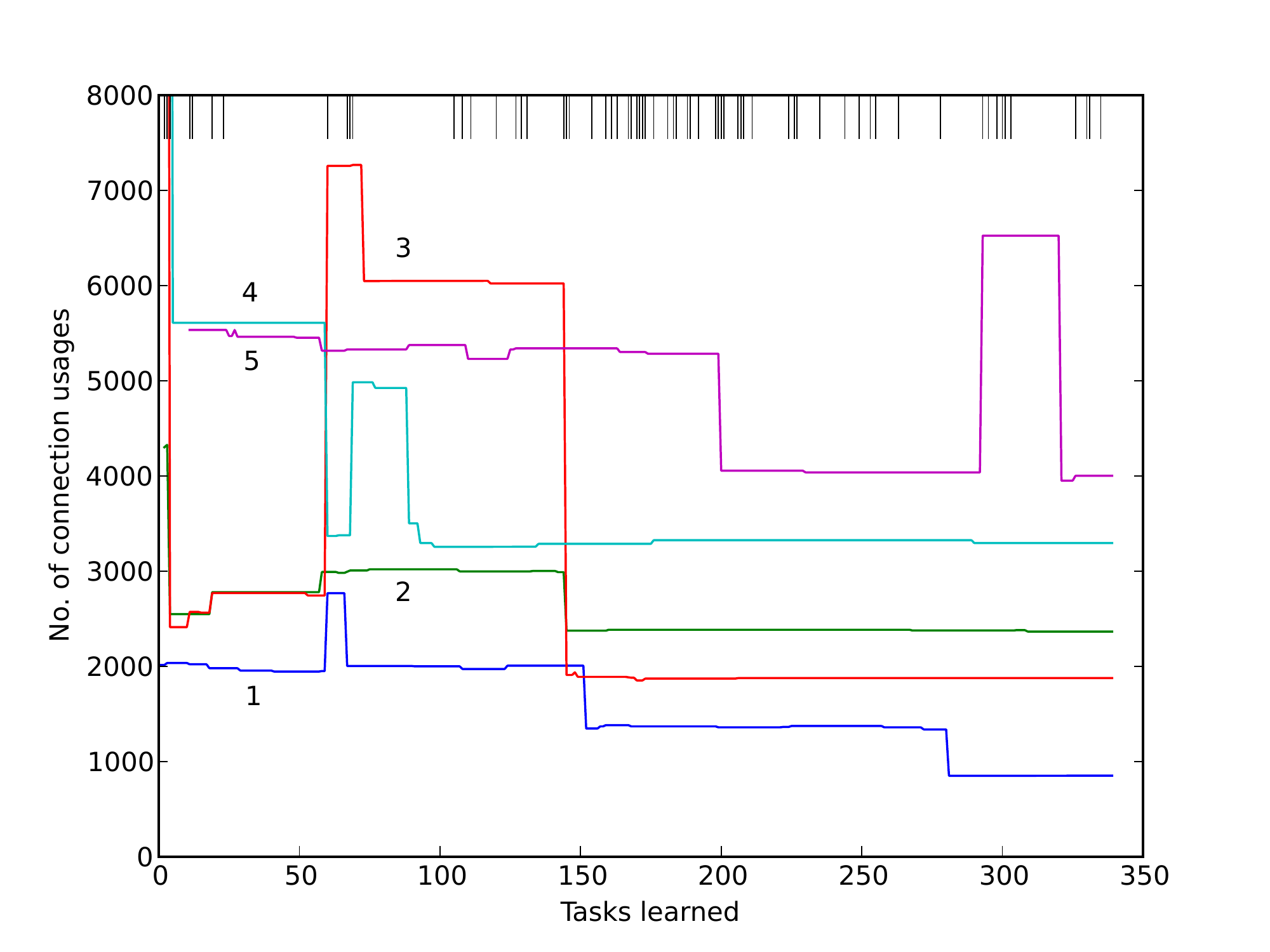}
  \caption{For the first five self-invented non-compression tasks, we plot the number of connection usages per task. In this run,  solutions to 340 self-generated tasks were learned. 67 of them were non-compression tasks (marked by small black lines at the top); the rest resulted in successful compressions of the SLIM RNN's weight matrix.  Over time, previously learned skills tend to require less and less computational resources, i.e., the SLIM RNN-based solver learns to speed up its solutions to previous self-invented tasks. Although some plot lines occasionally go up, this is compensated for by a decrease of connection usages for dozens of other tasks  (not shown here to prevent clutter).}
  \label{fig:tasksteps}
\end{figure*}

The network's internal states can be viewed as abstract summaries of its trajectories through the 
fovea environment and its parallel ``internal thoughts.'' The system invents more and more novel skills, each breaking
the generalization ability of its previous SLIM NN weight matrix, without forgetting previously learned skills. 
Within 8 hours on a 
standard PC, a SLIM RNN consisting of 20 WITAS,  with 4 neurons in each WITAS, invented 67 novel
action sequences guiding the fovea before halting. These 
varied in length, consuming up to 27 steps.
Over time the SLIM NN not only invented new skills to solve novel tasks, 
but also learned to speed up solutions to previously learned tasks, as shown in 
Fig.~\ref{fig:tasksteps}. For clarity, all figures presented here depict aspects of this same run, though results were consistent over many different runs.

\begin{figure*}[t]
  \centering
  \includegraphics[width=1.0\linewidth]{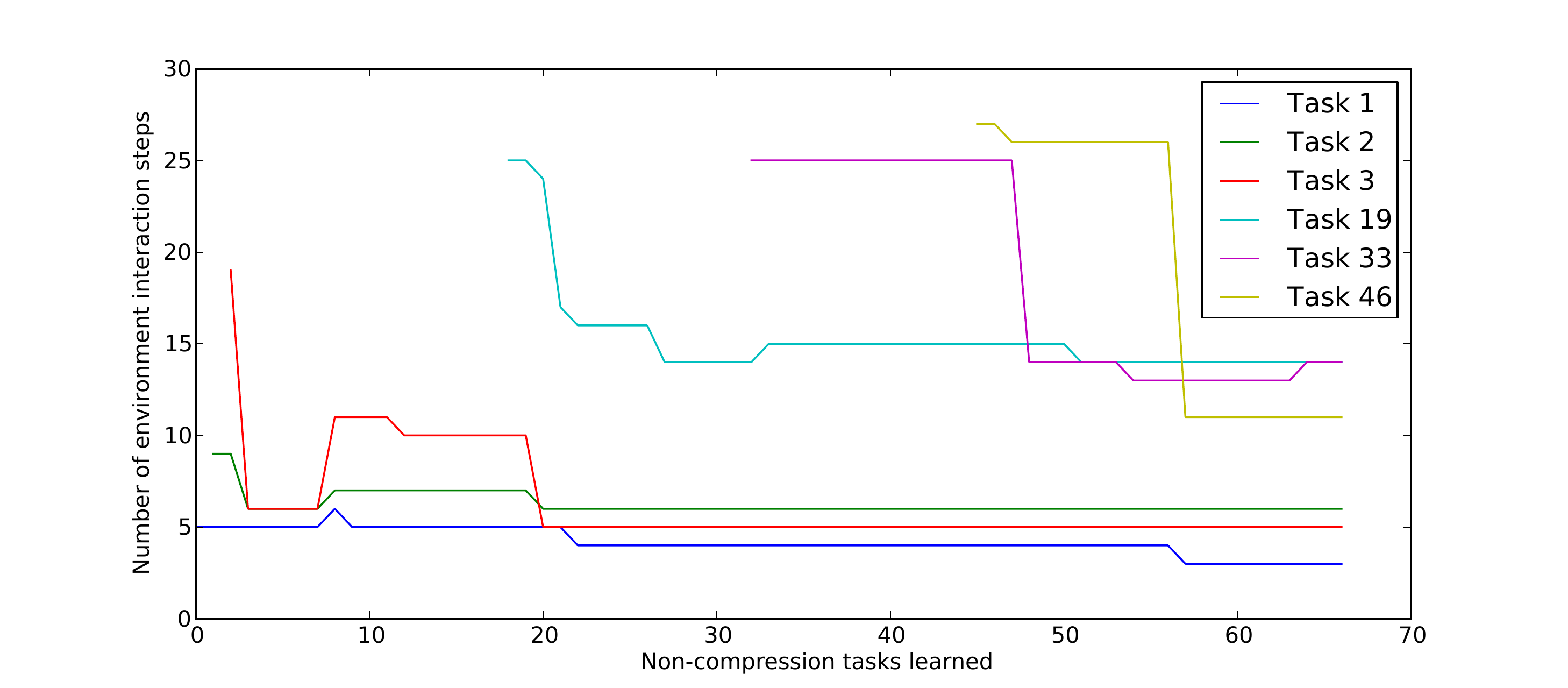}
  \caption{For only six selected tasks (to prevent clutter), we plot the number of interactions with the environment, over a run where 67 novel non-compression tasks were learned, besides numerous additional compression tasks ignored here. Here an interaction is a SLIM NN computation step that produces at least one non-zero output neuron activation. The total number of interactions cannot exceed the number of steps until the halt neuron is activated.}
  \label{fig:envsteps}
\end{figure*}

The SLIM NN also learns to reduce the interactions with the environment. Fig.~\ref{fig:envsteps} shows the number of interactions required to solve certain previously learned fovea control tasks. Here an ``interaction'' is a SLIM NN computation step that produces at least one non-zero output neuron activation. General trend over different tasks and runs: the interactions decrease over time. That is, the SLIM NN essentially learns to build internal representations of its interaction with the environment, due to  {\sc PowerPlay}'s continual built-in pressure to speed up and simplify and generalize.

\begin{figure*}[t]
  \centering
  \includegraphics[width=0.7\linewidth]{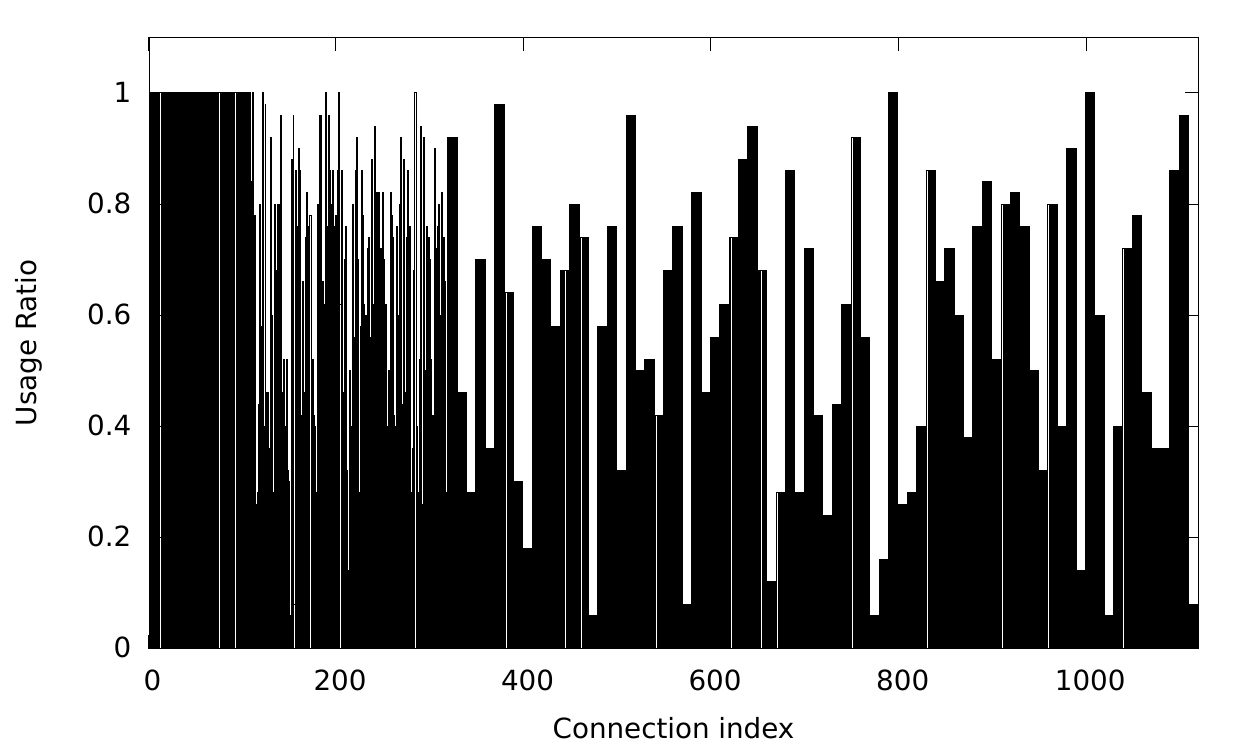}
  \caption{Connection usage ratios for all SLIM RNN connections after learning 227 out of the 340 total self-invented 
  tasks, 50 of them non-compression tasks forming the so-called task repertoire, the rest compression tasks. The \emph{usage ratio} on the $y$-axis is the number of repertoire tasks using the connection, divided by the number of repertoire tasks. This ratio is 1 for the first 110 connections, which are frequently used outgoing connections from task and online inputs. The network learns to better utilize its own architecture by using different connections for different tasks, thus reducing the number of connections with high usage ratio. Such modularization can help to speed up task search in later stages.}
  \label{fig:connusage}
\end{figure*}

The SLIM NN often uses partially overlapping subsets of connection weights for generating different 
self-invented trajectories. Fig.~\ref{fig:connusage} shows that not all connections are used for all tasks, and that  the connections used to solve individual tasks can become progressively more separated. In general, the variation in degree of separation depends on network parameters and environment.

\begin{figure*}[t]
  \centering
  \includegraphics[width=0.9\linewidth]{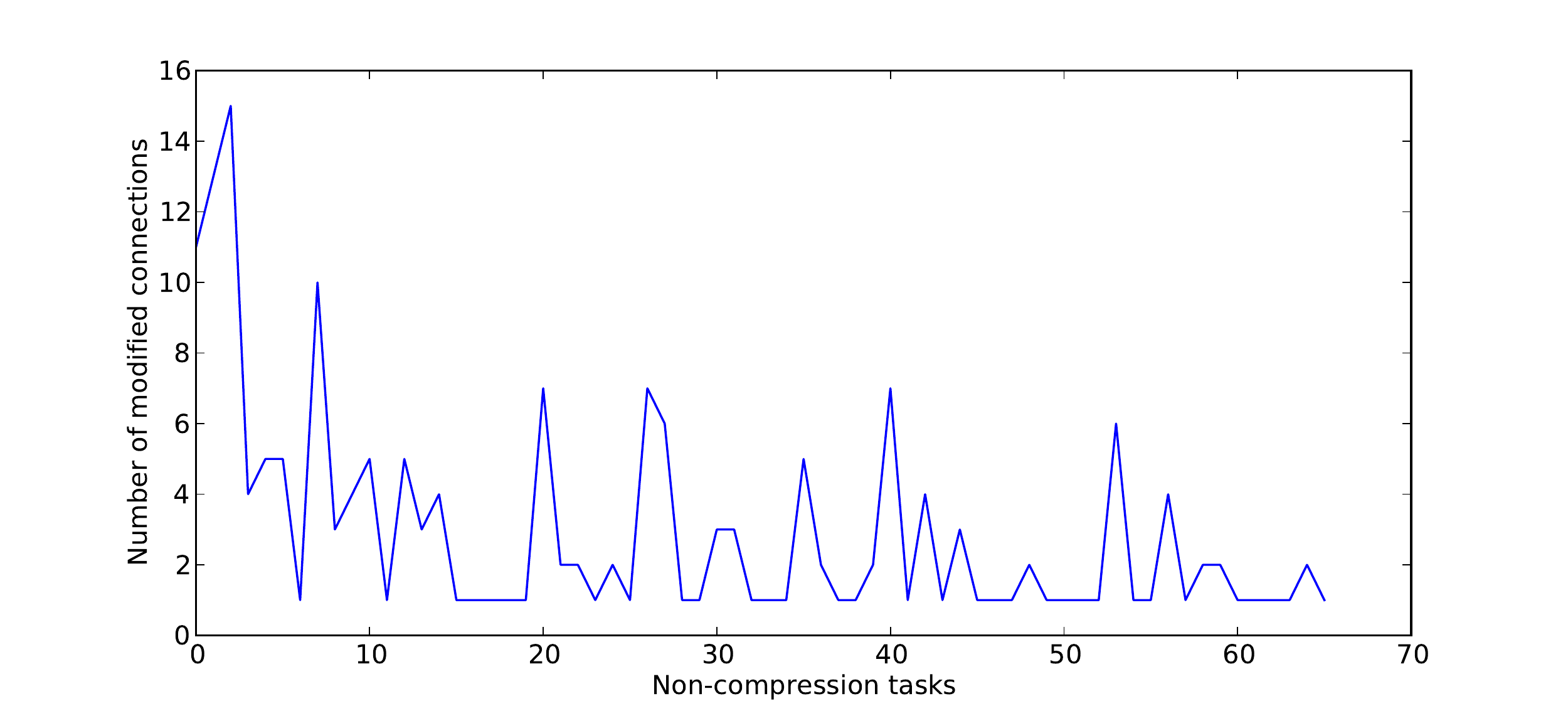}
  \caption{For each self-invented non-compression task, we plot the number of  modified SLIM NN weights needed to learn it without forgetting solutions to old tasks. During task search, the number of connections to modify is chosen randomly. Once the growing repertoire has reached a significant size, however, \emph{successfully} learned additional tasks tend to require few weight changes affecting few previous tasks (especially tasks with computationally expensive solutions). This is due to  {\sc PowerPlay}'s bias towards tasks that are fast to find and validate on the entire repertoire. See text for details.}
  \label{fig:changedwts}
\end{figure*}

As expected,  {\sc PowerPlay}-based SLIM NNs prefer to modify only few connections per novel task. Randomly choosing one to fifteen weight modifications per task, on average only $2.9$ weights were changed to invent a new skill---see Fig.~\ref{fig:changedwts}. Why? Because {\sc PowerPlay} is always going for the novel task that is fastest to find and validate, and fewer weight changes tend to affect fewer previously learned tasks; that is, less time is needed to re-validate performance on previous tasks.
In this way {\sc PowerPlay} avoids a naively expected slowdown linear in the number of tasks. Although the 
number of skills that must not be forgotten grows all the time,
the search time for new 
skills does not at all have to grow in proportion to the number of previously solved tasks.

As a consequence of its bias towards fast-to-validate solutions, the {\sc PowerPlay}-based SLIM NN automatically self-modularizes.  The SLIM RNN tested above had 1120 connections. Typically, 600 of them were used to solve a particular task, but on average less than three of them were changed. This means that for each newly invented task, the system re-uses a lot of previously acquired knowledge without modification. The truly novel aspects of the task and its solution often can be encoded within just a handful of bits. 

This type of self-modularization is more general than what can be found in traditional (non-inventive) modular reinforcement learning (RL) systems whose action sequences are chunked into macros to be re-used by higher-level macros, like in the options framework \cite{Precup:ICML98}, or in hierarchical RL \cite{Wiering:97ab}.  Since the SLIM RNN is a general computer, and its weights are its program, subsets of the weights can be viewed as sub-programs, and new sub-programs can be formed from old ones in essentially arbitrary computable ways, like in general incremental program search \cite{Schmidhuber:04oops}.

\section{Discussion and Outlook}
\label{creativity}

{\sc PowerPlay} for SLIM RNN represents a greedy implementation of central aspects of
the Formal Theory of Fun and Creativity
\cite{Schmidhuber:06cs,Schmidhuber:10ieeetamd}.
The setup permits  practically feasible, curious/creative agents
that learn hierarchically and modularly, using general computational problem solving architectures. 
Each new task invention either breaks the solver's present generalization ability, 
or compresses the solver, or speeds it up.

We can know precisely what is learned by {\sc PowerPlay}ing SLIM NN. The self-invented tasks are clearly defined by inputs and abstract internal outcomes / results. Human interpretation of the NN's weight changes, however, may be difficult, a bit like with a baby that generates new internal representations and skills or skill fragments during play. What is their ``meaning'' in the eyes of the parents, to whom the baby's internal state is a black box?  For example, in case of the fovea tasks the learner invents certain input-dependent movements as well as abstractions of trajectories in the environment (limited by its vocabulary of internal states). The RNN weights at any stage encode the agent's present (possibly limited) understanding of the environment and what can be done in it. 

 {\sc PowerPlay} has no problems with noisy inputs from the environment. However, a noisy version of an old, previously solved task must be considered as a new task, because in general we do not know what is noise and what is not. But over time {\sc PowerPlay} can automatically learn to generalize away the ``noise,'' eventually finding a compact solver that solves all ``noisy'' instances seen so far.

Our first experiments focused on developmental stages of purely creative systems, and did not involve any externally posed tasks yet. Future work will test the hypothesis that systems that have been running {\sc PowerPlay} for a while will be faster at solving many user-provided tasks than systems without such purely explorative components.  This hypothesis is inspired by babies who creatively seem to invent and learn many skills autonomously, which then helps them to learn additional teacher-defined external tasks. We intend to identify conditions under which such knowledge transfer can be expected.

\appendix
\section{Appendix: Implementation details}
\label{appendix}

The SLIM RNN used for Experiment 2 (fovea control) is constructed as follows:

Let the number of input, output and state neurons in the network be \emph{n\_input}, \emph{n\_output} and \emph{n\_state}, respectively. Let \emph{nb\_comp} = number of computation blocks each with \emph{block\_size} neurons. Thus there are \emph{nb\_comp}$\times$\emph{block\_size} computation neurons in the network.

The network is wired as follows. Each task input neuron is connected to \emph{nb\_comp} computation neurons at random. Each online input neuron is connected to \emph{nb\_comp}/10  neurons at random. Each internal state neuron receives connections from \emph{nb\_comp}/2 random computation neurons. The halt neuron recieves connections from \emph{nb\_comp}/2 random computation neurons. \emph{nb\_comp}$\times$\emph{n\_output} random computation neurons are connected to random output neurons. Each neuron in each computation block is randomly connected to \emph{block\_size} other computation neurons.

We used \emph{nb\_comp = 20}, \emph{block\_size = 4}, and \emph{n\_state = 3} with \emph{n\_input = 25} and \emph{n\_output = 8\emph} for the fovea control task. The \emph{halt-threshold} was set to 3, and the WITAS and fovea control \emph{win-threshold}s were set to $0.00001$. All connection weights were initialized to random values in $[-1,1]$. The cost of using a connection (consuming part of the \emph{time\_budget}) was set to $0.1$ for all connections. The mutation rule was as follows. For non-compression tasks, the network is first run using the new task inputs to check if the task can already be solved by generalization. If not, we randomly generate an integer number $m$ between 1 and 1/50th of all connections used during the unsuccessful run, and randomly change $m$ weights by adding to them a uniformly random number in $[-0.5,0.5]$. For compression tasks, we randomly generate a number $m$ between 1 and 1/50th of all connections used for any of the tasks in the current repertoire, and randomly modify $m$ of those connections.

The time budget fraction available to check whether a candidate task is not yet solvable by $s_{i-1}$ was chosen randomly between 0 and \emph{time\_budget}/2. For compression tasks, the sum of squared weights had to decrease by at least a factor of 1/1000 to be acceptable.

\section*{Acknowledgments} \label{ack} 
{\sc PowerPlay} \cite{Schmidhuber:11powerplay} and self-delimiting recurrent neural networks (SLIM RNN) \cite{Schmidhuber:12slimnn} 
were developed by  J.~Schmidhuber and implemented by R.K.~Srivastava and B.R.~Steunebrink.
We thank M.~Stollenga and N.E.~Toklu for their help with the
implementations.
This research was funded by the following EU projects: {IM-CLeVeR 
(FP7-ICT-IP-231722)} and {WAY (FP7-ICT-288551)}.

\bibliographystyle{plain}
\bibliography{bib}

\end{document}